\documentclass[10pt]{article}

\usepackage[parfill]{parskip}  
\usepackage{graphicx}
\usepackage{amsmath,amssymb,amsthm}
\usepackage{epstopdf}
\usepackage{xifthen}
\usepackage{enumerate}
\usepackage{enumitem}
\usepackage[colorlinks]{hyperref}
\usepackage{mathtools}
\usepackage[displaymath, mathlines]{lineno}
\usepackage{sectsty}
\sectionfont{\fontsize{10}{15}\selectfont}
\subsectionfont{\fontsize{10}{15}\selectfont}
\subsubsectionfont{\fontsize{10}{15}\selectfont}

\usepackage{hyperref}
\usepackage{tikz}
\usetikzlibrary{patterns,arrows,calc,decorations.pathmorphing}
\usetikzlibrary{matrix,chains,positioning,decorations.pathreplacing,arrows}
\usetikzlibrary{positioning,calc}
\usetikzlibrary{arrows,backgrounds}
\usetikzlibrary{angles,quotes}
\usetikzlibrary{cd}

\usepackage{algorithm}
\usepackage{algpseudocode}

\usepackage{rose}

\hypersetup{
    colorlinks=true,
    linkcolor=blue,
    filecolor=magenta,      
    urlcolor=cyan,
    pdftitle={Overleaf Example},
    pdfpagemode=FullScreen,
    citecolor = blue,
    }

\usepackage{subfigure}

\usepackage[style=alphabetic-verb,backend=bibtex,maxbibnames=99]{biblatex}

\usepackage{caption}

\usepackage{hyperref}
\usepackage{url}
\usepackage{rose}
\usepackage{todonotes}
\usepackage{graphicx,wrapfig}
\usepackage{caption}

\usepackage{algorithm}
\usepackage{algpseudocode}

\title{\LARGE Improved Representation Learning\\ Through Tensorized Autoencoders}
\author{\normalsize\textbf{	Pascal Esser\footnote{Equal contribution.}, \quad Satyaki Mukherjee$^*$,  \quad Mahalakshmi Sabanayagam$^*$,  \quad  Debarghya Ghoshdastidar}\\
\normalsize{Technical University of Munich} \\
\normalsize{School of Computation, Information and Technology}\\
\\
\normalsize{\textsc{\{esser, mukherjee, sabanaya, ghoshdas\}@cit.tum.de} }}

\begin{document}

\maketitle

\begin{abstract}
The central question in representation learning is what constitutes a good or meaningful representation. In this work we argue that if we consider data with inherent cluster structures, where clusters can be characterized through different means and covariances, those data structures should be represented in the embedding as well. While Autoencoders (AE) are widely used in practice for unsupervised representation learning, they do not fulfil the above condition on the embedding as they obtain a single representation of the data. To overcome this we propose a meta-algorithm that can be used to extend an arbitrary AE architecture to a tensorized version (TAE) that allows for learning cluster-specific embeddings while simultaneously learning the cluster assignment. For the linear setting we prove that TAE can recover the principle components of the different clusters in contrast to principle component of the entire data recovered by a standard AE. We validated this on planted models and for general, non-linear and convolutional AEs we empirically illustrate that tensorizing the AE is beneficial in clustering and de-noising tasks.
\end{abstract}

\section{INTRODUCTION AND MOTIVATION}\label{sec: introduction}

With the increasing use of very high dimensional data an important task is to find a good lower dimensional representation either to reduce noise in the data or to overcome the curse of dimensionality.
An obvious question is therefore, what is a \emph{good} representation and how can we guarantee that a given algorithm obtains it?
Conceptually a latent representation should preserve certain structures of the data, while removing noise dimensions. Naturally there are different perspectives on what could constitute an important structure of the given data, such as an underlying topology or clustering structures. In this work we consider the latter from a statistical perspective and more specifically from the perspective of its covariance structure and latent representation. Suppose our dataset consists of $k$ clusters that each have some inherent structure. Then we would want those separate structures to be again represented in the latent space. 
One of the most common approaches in deep learning for obtaining latent representations are Autoencoders (AE) \cite{Kramer1991AIChE_AE}.
Formally we write a general AE as the following optimization problem
\begin{align}\label{eq: general AE}
    \min_{\Phi,\Psi}\norm{{\bX} - f_\Phi\left(g_\Psi\left({\bX}\right)\right)} + \lambda*\mathrm{regularizer},
\end{align}
where ${\bX}\in\bbR^{d\times n}$ is the \emph{centred} data input matrix consisting of $n$ samples of dimension $d$  and $\widehat{\bX} =f_\Phi\left(g_\Psi\left({\bX}\right)\right) $ the reconstruction. Let $g_\Psi( \cdot )$ be the \emph{encoder function}, parameterized by $\Psi$ mapping from the data dimension $d$ onto the latent dimension $h$ and $f_\Phi( \cdot )$ the \emph{decoder function} parameterized by $\Phi$ mapping back to $d$. $\lambda$ determines the strength of the regularizer, that is chosen depending on the task or desired properties of the parameters.

\begin{figure}
    \centering
    \includegraphics[scale=.5]{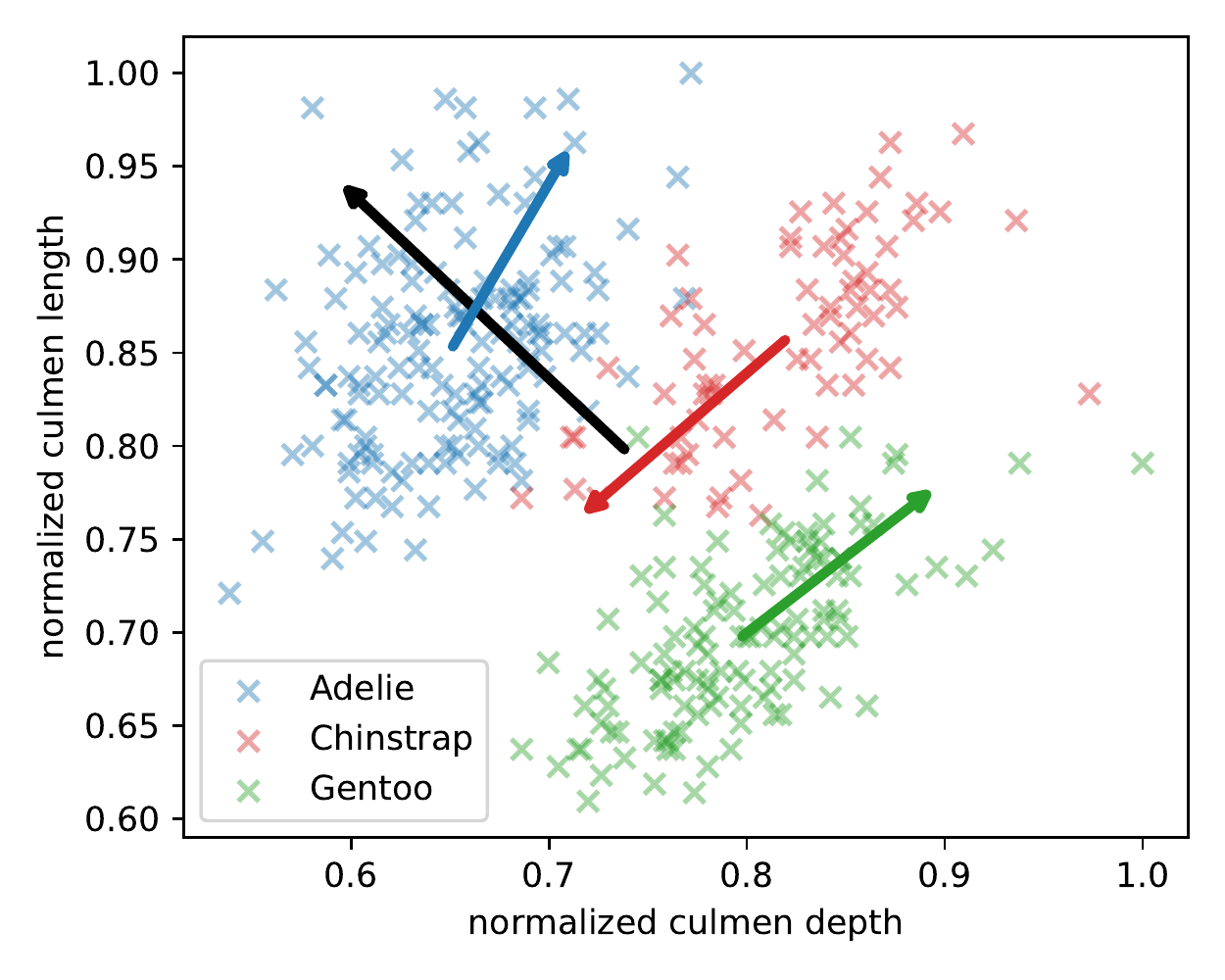}
    \caption{Illustration simpson's paradox in the `penguin dataset' \cite{Penguins}. The three clusters and their first principal component are plotted in red, blue and green respectively and the principal direction for the full dataset in black.}
    \label{fig:pca}
\end{figure}

It has been shown that in the linear setting the encoder recovers the principal components of the full dataset \cite{Kunin20219pmlr_AE_losss_landscape}. However this one representation does not necessarily capture the underlying cluster structure well as illustrated in Figure~\ref{fig:pca}.
A classical example of this is the so called \emph{simpson's paradox} \cite{Simpson1951stat}, a known phenomenon in statistics where the trend in clusters does not align with the trend that appears in the full dataset, often observed in social-science and medical-science statistics \cite{Clifford1982,Holt2016}.
Figure~\ref{fig:pca} shows that one representation for all the data points (shown in black) does not capture the structures of individual clusters well (shown by the colored arrows). To be able to model such structures we therefore introduce a modified AE architecture we term \emph{Tensorized Autoencoders (TAE)} that, in the linear setting, provably recovers the principal directions of \emph{each cluster} while jointly learning the cluster assignment.
This new AE architecture considers a single AE for each cluster allowing us to learn distinct cluster representations. Important to note that while this increases the number of parameters, the representation still remains the same dimension as before. In particular this still experimentally performs better than a single autoencoder with $d \times k$ encoding dimension. Formally we change \eqref{eq: general AE} to the following 
\begin{align}\label{eq: general TAE}
    \min_{\{\Phi_j, \Psi_j\}^k_{j=1},\bS}\sum^n_{i=1} & \sum^k_{j = 1}\bS_{j,i}\left[\norm{(\bX_i-\bC_j) - f_{\Phi_j}\left(g_{\Psi_j}\left({\bX}_i-\bC_j\right)\right)}  + \lambda*\mathrm{regularizer}_j\right],
\end{align}
where $\bS$ is a $k \times n$ matrix, such that $\bS_{j,i}$ is the probability that data point $i$ belongs to class $j$, $g_{\Psi_j}( \cdot )$ and $f_{\Phi_j}( \cdot )$ are the \emph{encoder }and the \emph{decoder} functions respectively specific to points in class $j$, and $C_j$ is a parameter centering the data passed to each autoencoder specific to class $j$.
Specifically a $k$-means \cite{Macqueen67somemethods} regularizer is considered to enforce a cluster friendly structure in the latent space (similar to the regularizer proposed in \cite{Bo2017k_means_friendly}).

To further illustrate the importance of the new formulation \eqref{eq: general TAE}, that provides cluster specific representations, we look at two important representations learning downstream tasks.
\begin{figure*}
    \centering
    \includegraphics[trim={4cm 0 3.5cm 0},clip,width= \textwidth]{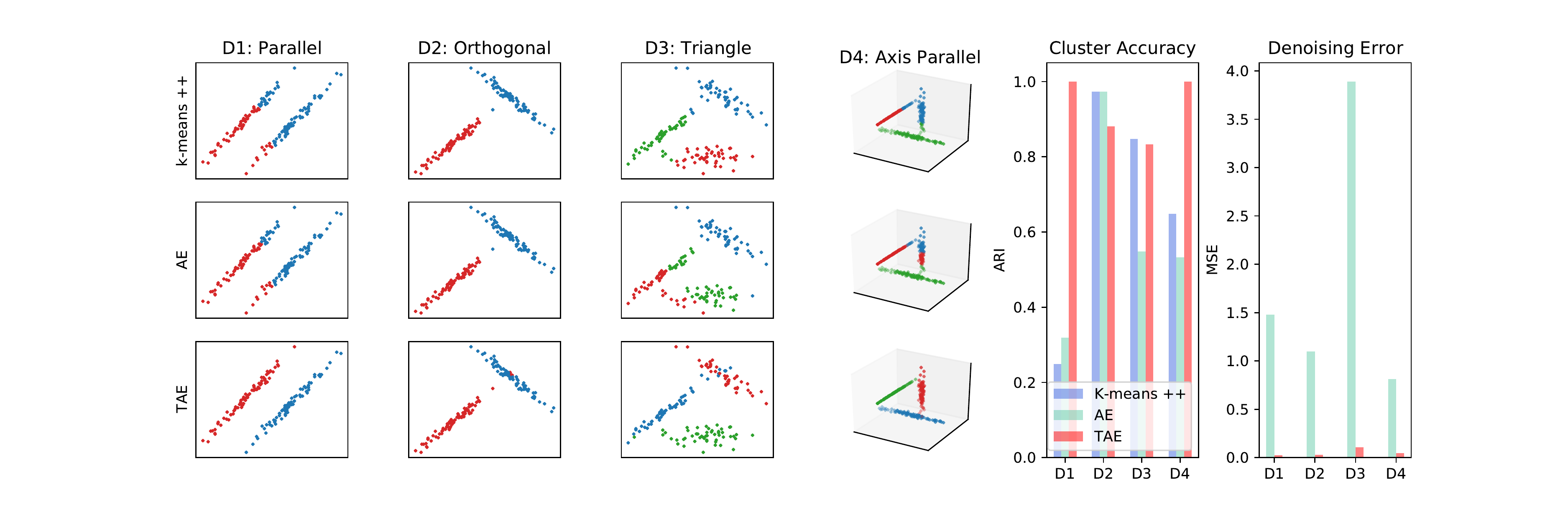}
    \caption{Illustration for cases where separate embedding for clusters are beneficial. 
    \textbf{Left grid.} Plotted is the true data in $\bbR^2$. Top row: Clustering obtained from $k$-means ++ on the original features. Middle row: $k$-means ++ on the embedding obtained by a standard AE 
    Bottom row: $k$-means ++ on the proposed TAE.
    \textbf{Right plot.} Alignment of clustering with true labels measured by ARI \cite{ARI}.
    }
    \label{fig: toy data}
\end{figure*}

\textbf{Clustering.} The main goal of clustering is to group similar objects into the same class in an unsupervised setting. While this problem has been extensively studied in traditional machine learning the time complexity significantly increases with high dimensional data and therefore existing works focus on projecting data into low-dimensional spaces and then cluster the embedded representations \cite{Volker2003_NeurIPS_feature_selection,Fei2014AAAI_graph_representation_clustering,Zhangyang2016SIAM_Architecture_clustering}. 
Several methods have been developed that use deep unsupervised models to learn representations with a clustering focus that simultaneously learns feature representations and cluster assignments using deep neural networks \cite{Xie2016ICML_deep_embedding,Kamran2017_ICCV_Joint_AE,Zhangyang2016SIAM_Architecture_clustering,Xie2015IntegratingIC,Zhangyang2015ICAI_Discriminative_Clustering}. However all of these algorithms learn a single representation for the full dataset.
Assuming we use an AE for learning the representation we formally consider \eqref{eq: general AE} and perform the clustering on $g_\Psi\left({\bX}\right)$ instead of $\bX$. 
Now the question is, is one representation of the data sufficient?
We investigate this question by considering different planted datasets as shown in Figure~\ref{fig: toy data}, where we compare the clustering obtained from $k$-means ++ \cite{David2007kmeans} on the original features, the embedding obtained by a standard AE and the proposed TAE.
The advantage of TAE is that it captures the \emph{directions} of the clusters whereas simple AE mis-classifies some points in clusters that are close in euclidean space.
For those datasets the distance function is inherently linked to the shape of the clusters.
We extend this analysis to real world data and more complex AE architectures in Section~\ref{sec: experiments}.

\textbf{De-noising.} Consider an image corrupted by noise, the task is to remove the noise from the image.  In this setting a good representation is one that only returns the true data structure and removes the noise. Here we are not directly interested in the embedding but only in the obtained reconstruction of the decoder. However we again conjecture that having separate embeddings for each cluster is beneficial as it allows to learn more cluster specific representations.

We consider a De-Noising AE (DAE), formally defined \cite{Antoni2005SIAP_denoising_survay,Cho2013pmlr_DAE}  by considering $ \bX + \varepsilon$  as input in Eq~\ref{eq: general AE} where $\varepsilon$ is an additive noise term and the goal is to remove the noise from the data.  
An additional advantage in the denosing setting is that the cluster structure obtained by the TAE does not have to match the one imposed by the ground truth. This is beneficial in cases where there is not a unique way to cluster given data (e.g. when clustering cars, one might group by color or by type). We see this advantage empirically illustrated in Figure~\ref{fig: toy data}, right, where we see that the MSE of the simple AE is consistently significantly higher than for the TAE. 
We further validate this intuition empirically in  Section~\ref{sec: experiments}.

\textbf{Contributions}
This paper provides the following main contributions.
(a) To learn different representations for each cluster in the data we formally propose tensorized autoencoders with $k$-means regularisation in section~\ref{sec: strict cost}, that simultaneously learns the embedding and cluster assignments. 
(b) In section~\ref{sec: parameterization at optimum} we prove that this architecture with linear encoder and decoder recovers the $k$ leading eigenvectors of the different clusters instead of the eigenvectors of the whole data-set as done by standard linear AEs. 
(c) Empirically we demonstrate in section~\ref{sec: experiments} that the general concept can be extended efficiently to the non-linear setting as well and TAE perform well in clustering and de-noising tasks on real data.
(d) Finally in section~\ref{sec: EM} we show how TAE are connected to Expectation maximization.
We provide related works and concluding remarks in sections \ref{sec: related work}--\ref{sec: conclusion} and all the proofs in Appendix.

\textbf{Notation.} We denote vectors as lowercase bold, $\ba$, matrices as uppercase bold, $\bA$ and tensors calligraphic as $\mathcal{A}$. Let $\norm{~\cdot~}^2$ be the standard square norm and a function $f$ parameterized by $\Theta$ as $f_\Theta$.
Furthermore let $\1_n$ be the all ones vector of size $n$ and $\bbI_n$ the $n\times n$ identity matrix.

\section{ANALYSIS OF TENSORIZED LINEAR AE WITH K-MEANS REGULARIZER}\label{sec: strict cost}

To get a better understanding of the behaviour of the proposed architecture in \eqref{eq: general TAE}, we consider a simple \emph{linear-AE} as this allows us to analytically derive the optimal parameterization. We extend the analysis empirically to a non-linear setting in section~\ref{sec: experiments}.

\subsection{Formal Setup}
For simplicity consider the clustering setting where $\widetilde{\bX} := \bX\in\bbR^{n\times d}$, with $n$ being the number of data-points and $d$ the feature dimension. Note that the analysis extends directly to the de-noising setting as well. 
For a two layer linear AE, let $\mathcal{U} \in \bbR^{k \times h \times d}$ and $\mathcal{V} \in \bbR^{k \times d \times h}$ be the encoding and decoding tensor respectively. Then for each $1 \leq j \leq k$, let
$\bU_{j}\in\bbR^{h \times d}$, be the  \emph{encoding} matrix by taking the appropriate slice of the tensor $\mathcal{U}$.  Essentially $\bU_j$ corresponds to the encoding function of the $j$'th cluster. We assume $\bU_{j}$ is a projection matrix i.e. $\bU_{j}\bU_{j}^T = \bbI_k$. Similarly define $\bV_{j}^{d \times k}$ from the \emph{decoder} $\bV$ (injection matrix). Finally let  $\bC_{j}$ be the cluster centers and $\lambda$ be the weight assigned to the regularizer. We treat it as a hyperparameter. From there we define the loss function as
 \begin{align}\label{eq: relaxed cost}
     \loss_\lambda(\bX) := &\sum^n_{i=1} \sum^k_{j = 1}\bS_{j,i}\Big[\norm{\left(\bX_i - \bC_j\right) - \bV_j\bU_j\left(\bX_i - \bC_j\right)}^2 ~ + ~ \lambda \norm{\bU_j\left(\bX_i - \bC_j\right)}^2\Big], \nonumber\\
     \text{s.t. ~ }& \1_k^T\bS = \1_n^T,\ \bS_{j,i} \geq 0,
 \end{align}
where we define $\bS$ to be a $k \times n$ matrix, such that $\bS_{j,i}$ is the probability that datapoint $i$ belongs to class $j$. To ensure that the entries can be interpreted as probabilities we impose the above stated constraints. To further illuminate the intuition behind $\bS$  consider a dataset $\{\bX_i\}_{i=1}^n$ where associated to each datapoint $\bX_i$ there are probabilities $\{\bS_{j,i}\}_{j=1}^k$ corresponding to how certain we are that $\bX_i$ is sampled from the true distribution. These $\{\bS_{j,i}\}$ are latent variables that we learn from the dataset. 

\subsection{Parameterization at optimum}\label{sec: parameterization at optimum}

For the linear setting we derive the parameterization at the optimum but for reference we first recall the optimum of a standard linear AE:
\begin{theorem}[Parameterization at Optimal for Linear AE]\label{Th: linear AE optimal}
\cite{BALDI1989NN} showed that linear AE without regurlarization finds solutions in the principal component spanning subspace, but the individual components and corresponding
eigenvalues cannot be recovered.
\cite{Kunin20219pmlr_AE_losss_landscape} show that $l_2$ regularization reduces the symmetry solutions to the group of orthogonal transformations.
Finally \cite{Bao2020Neurips_AE_PC} show that non-uniform $l_2$ regularization allows linear AE to recover ordered, axis-aligned principal components.
\end{theorem}
From this we note that we only learn a single representation for the data and therefore cannot capture the underlying cluster structures.
To give the intuition of this cost, before we characterize the optimal parameters of a linear TAE, lets consider a single datapoint $\bX_i$ assigned to cluster $\bC_j$ and its cost is,
\begin{align*}
    \bS_{j,i}\left[\norm{\left(\bX_i - \bC_j\right) - \bV_j\bU_j\left(\bX_i - \bC_j\right)}^2 + \lambda \norm{\bU_j\left(\bX_i - \bC_j\right)}^2\right].
\end{align*}
Ignoring $\bC_j$, the cost is simply the cost of $\bX_i$ in AE weighted by the probability of it belonging to cluster $j$ (i.e. $\bS_{j,i}$). With this intuition, we characterize the optimal parameters of a linear TAE in the following theorem.

\begin{theorem}[Parameterization at Optimal for TAE]\label{Th: optimal relaxed}
For $0 < \lambda \leq 1$, optimizing Eq.~\ref{eq: relaxed cost} results in the parameters at the optimum satisfying the following:
\begin{enumerate}
    % \item[i)] 
    \item
    \textbf{Class Assignment.} While in Eq.~\ref{eq: relaxed cost} we define $\bS_{j,i}$ as the probability that $\bX_i$ belongs to class $j$ at the optimal $\bS_{j,i} = 1 \textit{ or } 0$ and therefore converges to a strict class assignment.
    
    % \item[ii)] 
    \item
    \textbf{Centers.} $\bC_j$ at optimum naturally satisfies the condition 
    \begin{align*}
    \bC_j = \frac{\sum_i \bS_{j,i}\bX_i}{\sum_i \bS_{j,i}}.
    \end{align*}

    % \item[iii)]
    \item
    \textbf{Encoding / Decoding (learned weights).}  We first show that $\bV_j^T = \bU_j$, and define 
    \[ 
    \hat{\mathbf{\Sigma}}_j: =\sum^n_{i=1}\bS_{j,i}\left(\bX_i - \bC_j\right)\left(\bX_i - \bC_j\right)^T,
    \] 
    then the encoding corresponds to the top $k$ eigenvectors of $\hat{\Sigma}_j$.
\end{enumerate}
\end{theorem}
  
At the above values for the parameters, $\bC_j$ and $\hat{\mathbf{\Sigma}}_j$ acts as estimates for the means and covariances for each specific class respectively. Thus assuming that $\bS$ gives reasonable cluster assignments, $\bU_j$ and $\bV_j$ combined gives the principal components of each cluster.  While points ii) and iii) follow directly by deriving the parameterization at the optimal and we give the full derivation in the supplementary material, we give a short intuition on i) here. First we note that as per our current definition $\loss$ is linear in $\bS$. Thus the global optimum of the loss with respect to the aforementioned linear conditions on $\bS$ must be at some vertex of the convex polytope defined by the conditions. Since these conditions are $\1_k^T\bS = \1_n^T$ and $\bS_{j,i} \geq 0$, at any of the vertices of the corresponding polytope we have that $\bS_{j,i} = 1 \textit{ or } 0.$ This combined with the fact that at global optimum $\bC_j$ satisfies the condition $\bC_j = \frac{\sum_i \bS_{j,i}\bX_i}{\sum_i \bS_{j,i}}$ implies that the global optimum of the loss $\loss$ in this expanded space is precisely same as that of the cost in the strict case we discuss in remark~\ref{rem: strict cost}. This is important as it shows that even though we allow the optimization using gradient decent the parameterization at the optimal assigns each datapoint to one AE.
  
Let us compare Theorem~\ref{Th: optimal relaxed} to the general notion proposed in the introduction: We would like an approach to obtain a separate meaningful representation (in the sense of recovering principal directions) for each cluster structure without having prior knowledge of which cluster a given datapoint belongs to.  Theorem~\ref{Th: optimal relaxed} shows that the proposed tensorized AE (Eq.~\ref{eq: relaxed cost}) fulfills those requirements as TAE recover the top $k$ eigenvectors for each cluster separately.

  \begin{remark}[Strict cost function]\label{rem: strict cost}
  Since we showed in Theorem~\ref{Th: optimal relaxed}  that  Eq.~\ref{eq: relaxed cost} converges to a strict class assignment we can alternatively also define the loss function directly with strict class assignments as follows:
  Let $\overline{\bX}^{S(i)}$ be the center of all data-points which $\bX_i$ belongs to. We define the loss function with strict class assignments as
  \begin{align*}
\min_S \min_{\bV_{S(i)},\bU_{S(i)}} \min_{C_{S(i)}} \sum_{i=1}^n \norm{\big(\bX_i - \overline{\bX}^{S(i)}\big) - \bV_{S(i)}\bU_{S(i)}\big(\bX_i - \overline{\bX}^{S(i)}\big)}^2+ \lambda\norm{\bU_{S(i)}\bX_i - \bC_{S(i)}}^2
\end{align*}
Similar to Theorem~\ref{Th: optimal relaxed} we again characterize the parameters at the optimal and most importantly note that with 
\begin{align*}
      \hat{\Sigma}_j = \frac{\sum_{i : S(i) = j} \big(\bX_i - \overline{\bX}^j\big)\big(\bX_i - \overline{\bX}^j\big)^T}{|\{i : S(i) = j\}|},
\end{align*}
as long as $\lambda < 1$, the optimal projection $U_j$ for the above cost is exactly the top $k$ eigenvectors of $\hat{\Sigma}_j$.
  \end{remark}

\subsection{Optimization}\label{sec: training}

While in the previous section we discussed the optimum that is obtained by solving the optimization problem in Eq.~\ref{eq: relaxed cost} we now look at how to practically train the tensorized AE. 
The general steps for learning the encoder and decoder are summarized as follows, where step 2 and 3 are repeated until convergence.
\begin{enumerate}
    \item \textbf{Initialize} weights and cluster assignments according to $k$-means ++ \cite{David2007kmeans}\footnote{Note on the initialization: while in this algorithm we consider a $k$-means++ in cases where we have access to some labels (e.g. in a semi-supervised setting) this can be replaced by considering random points with given labels for each cluster as initializations.}.
    \item \textbf{Update the weights} for the encoder and decoder (using e.g. a GD step).
    \item \textbf{Update the class assignment} $\bS$. To do so we consider a number of different options: 
    
    \emph{Option 1:} using a GD step under constraints ~$\1_k^T\bS = \1_n^T, ~ \bS_{j,i} \geq 0$.
    
  \emph{Option 2:} using an un-constrained GD step and project $\bS$ back onto the constraints.
  
      \emph{Option 3:} using a Lloyd’s step\footnote{Here the Lyod's step solves the linear problem on $\bS$ assuming all other parameters are fixed.} on a strict class-assignment.
\end{enumerate}
The choice of options in step 3 mostly depends on the framework of implementation. As noted, $\bS$ as defined in Eq.~\ref{eq: relaxed cost}  allows us to perform gradient updates on the class assignments. The advantage is that since $\bS$ is not defined binary, frameworks such as CVXPY \cite{diamond2016cvxpy} or Keras \cite{chollet2015keras} can be used to perform a constrained gradient steps, however in popular frameworks such as PyTorch \cite{Pytorch}, that as of the time of writing this paper do not directly support constraint optimization \emph{Option 3} can be used.

While the general goal of the proposed setup is to learn good data representations, the exact train and test steps depend on the downstream task. In this work there are two main settings. 
For instance in \emph{clustering} we directly apply the above steps and jointly learn the clustering and embedding. While in a simple AE we would have to apply a clustering algorithm onto the latent representations, in TAE the above steps directly provide the clusters. While for the linear case we prove that $\bS_{ij}$ is binary, in practice, especially with more complex networks one has to compute $\argmax_j\bS_{ij}$ to determine the class assignment.
On the other hand in \emph{de-nosing} we again jointly learn the embedding and cluster assignment but only train on the train set and in a second step use the learned TAE to de-noise images in the test set. While for a simple AE we can directly pass test data, with a TAE we use the approach presented in section~\ref{sec: new data} that allows us to assign the new datapoint to the appropriate AE to process it.
This same setup could be used for tasks such as super resolution or inpainting.
Importantly, while the results from Theorem~\ref{Th: optimal relaxed} hold only for the exact linear formulation in  Eq.~\ref{eq: relaxed cost}, the main idea and training steps can be extended to arbitrary encoding and decoding functions which we illustrate empirically in section~\ref{sec: experiments}.

\subsection{Test on new data}\label{sec: new data}

In the context of De-noising and validating clustering on an independent test dataset or other downstream tasks, we generally need a way of encoding and decoding a new or test datapoint. 
In contrast to AE, where one simply passes any new data through the trained network, TAE additionally decides on the latent variable $\bS$.
Note then that essentially in our model the actual parameters are $\bC_j, \bU_j, \bV_j$ whereas $\bS$ is simply an encoding of our confidence of what the latent variable or label is for each $\bX_i$. Thus to run the TAE on a new datapoint, we have to first estimate this latent variable.

Following this idea, let $\bs \in \bbR^k$, with its $j$'th coordinate being $\bs_j$.\footnote{Note then that this is a linear problem in the variables $\bs_j$ and thus its solution is at some vertex of the bounding polytope, i.e. there is some $\hat{j}$ such that $\hat{\bs}_{\hat{j}} = 1$. Thus the label assigned to $\widetilde\bX_i$ is this $j$. } Then given a new datapoint $\widetilde\bX_i$ and given the trained parameters $\bC_j, \bU_j, \bV_j$, we first find $\hat{\bs}$ such that 
\begin{align*}
    \hat{\bs} = \argmin_{\1_k^T\bs = 1; \bs_j \geq 0}  \sum^k_{j = 1} \bs_j \left[\norm{(\widetilde\bX_i - \bC_j) - \bV_j\bU_j(\widetilde\bX_i - \bC_j)}^2+ \lambda \norm{\bU_j(\widetilde\bX_i - \bC_j)}^2\right].
\end{align*}
We use this setting for example for a de-noising task such that the de-noised reconstruction of $\widetilde\bX_i$ would be given by $(\bbI-\bV_{\hat{j}}\bU_{\hat{j}})(\widetilde\bX_i-\bC_{\hat{j}}) + \bC_{\hat{j}}$.

We consider the following two cases to present a heuristic that the step in general does not worsen the prediction. On the one hand if \emph{clusters are well separated}, this implies that the separate AEs are quite different and therefore a wrong assignment would be significant. However well separated clusters also implies that the assignment of a new point is with high probability correct. On the other hand if the \emph{clusters are not well separated} then the point might be assigned to the wrong cluster more easily, however similar clusters also implies similar AEs so even if we assign a new point to the wrong AE the reconstruction is still close to the one from the correct AE.

\section{EXPERIMENTS WITH NON-LINEAR AND CONVOLUTIONAL NETWORKS}\label{sec: experiments}

The above linear setting allows us to perform a thorough analysis of the model and prove that the parameterization at the optimal fulfills the desired criteria of learning a meaningful representation for each cluster instead of one for the whole dataset. However in an applied setting we would like to take advantage of the expressive power of more complex, non-linear encoders and decoders as well. Therefore we  first discuss how tensorized AE can be extended to a more general setting and additionally empirically validate its performance.

All further details on the implementation and experiments are provided in the supplementary material.
\subsection{Extension to Arbitrary AE Architectures}

Given a single datapoint $x$ and some specific architecture, let $\loss_{\{\Phi_j, \Psi_j\}^k_{j=1}}(x)$  and $g_{\Psi_j}( \cdot )$ be the corresponding loss function and encoder parameterized by $\Psi_j$ respectively. Then the corresponding tensorized autoencoder is generated by first considering $k$ many independent copies of the above AE $\{g_{\Psi_j}( \cdot )\}_{j=1}^k.$ Then the tensorized loss is defined by 
\begin{align*}
    \loss(X) := \sum^n_{i=1} \sum^k_{j = 1}\bS_{j,i}\left[\loss_{\Phi_j, \Psi_j}\left(\bX_i - \bC_j\right) + \lambda \norm{g_{\Psi_j}\left(\bX_i - \bC_j\right)}^2\right],
\end{align*} where $\bC_j$ is defined as $\bC_j = \frac{\sum_i \bS_{j,i}\bX_i}{\sum_i \bS_{j,i}}$. 
While a more involved analysis is required to prove the exact latent representation in this case, 
%While for those cases we can not as easily prove what exact representation is recovered, 
the overall idea presented in \eqref{eq: general TAE}.
Again for the training we follow the steps presented in Section~\ref{sec: training}.

\subsection{Clustering on Real Data}
While on the toy data in Figure~\ref{fig: toy data} we clearly observed how the analyzed approaches deal with different datastructures, we extend the analysis to real data and also more complex AE architectures. 

For comparability of the number of parameters we furthermore compare \emph{AE 1} with latent dimension $d$ and \emph{AE 2} with latent dimensions $d \times k$ in Figure~\ref{fig:real clustering}. We note that their performance on the clustering task is very comparable and we therefore conclude that the performance of the TAE is not a direct result of the increased number of parameters but is achieved due to the different architecture.

\begin{figure}[t]
    \centering
    \includegraphics[width = 0.6\textwidth]{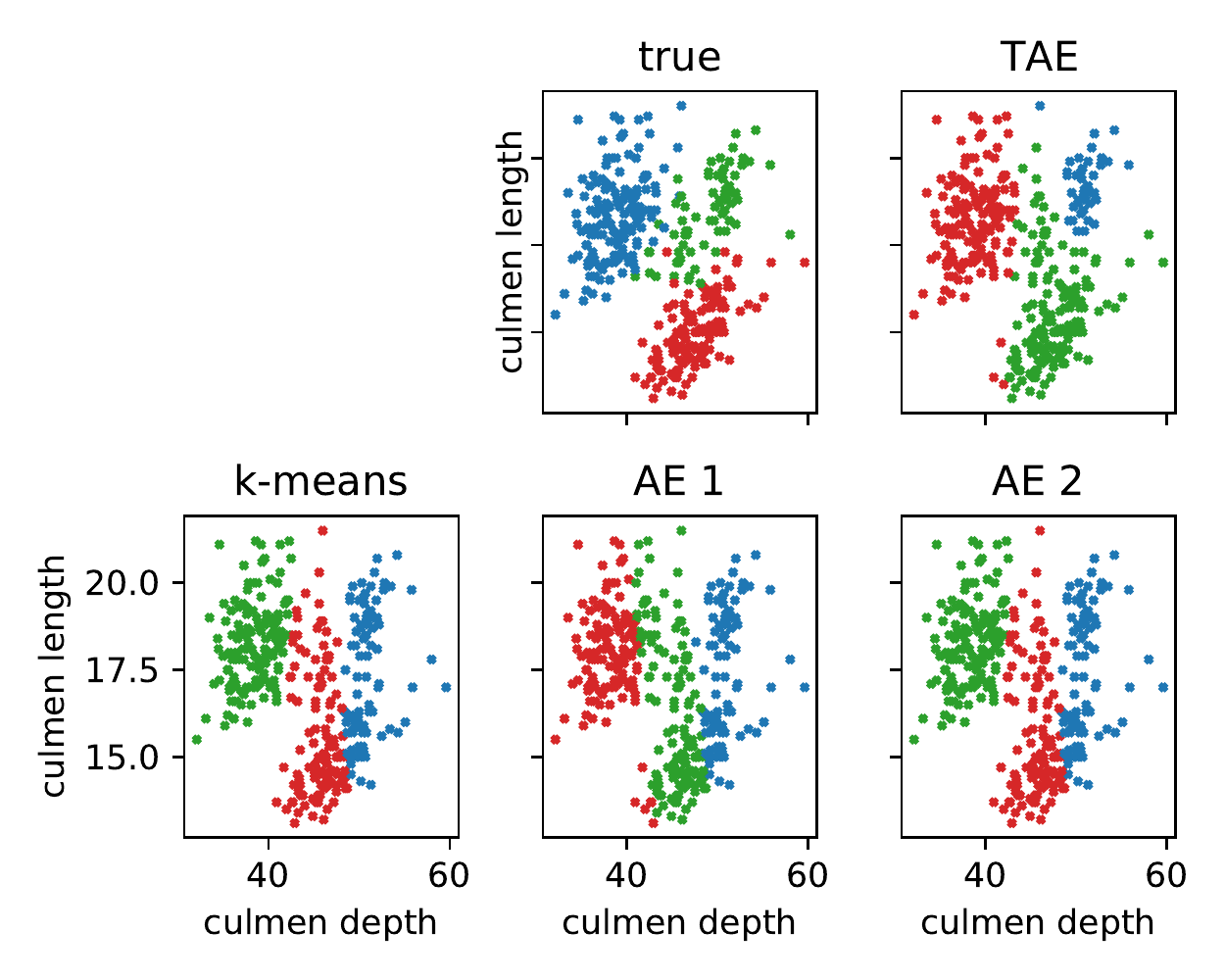}
    \caption{Illustration of the final clustering obtained for the \emph{penguin dataset, two features} by the different algorithms.}
    \label{fig:penugin scatter}
\end{figure}

\begin{figure}[t]
    \centering
    \includegraphics[width = 0.6\textwidth]{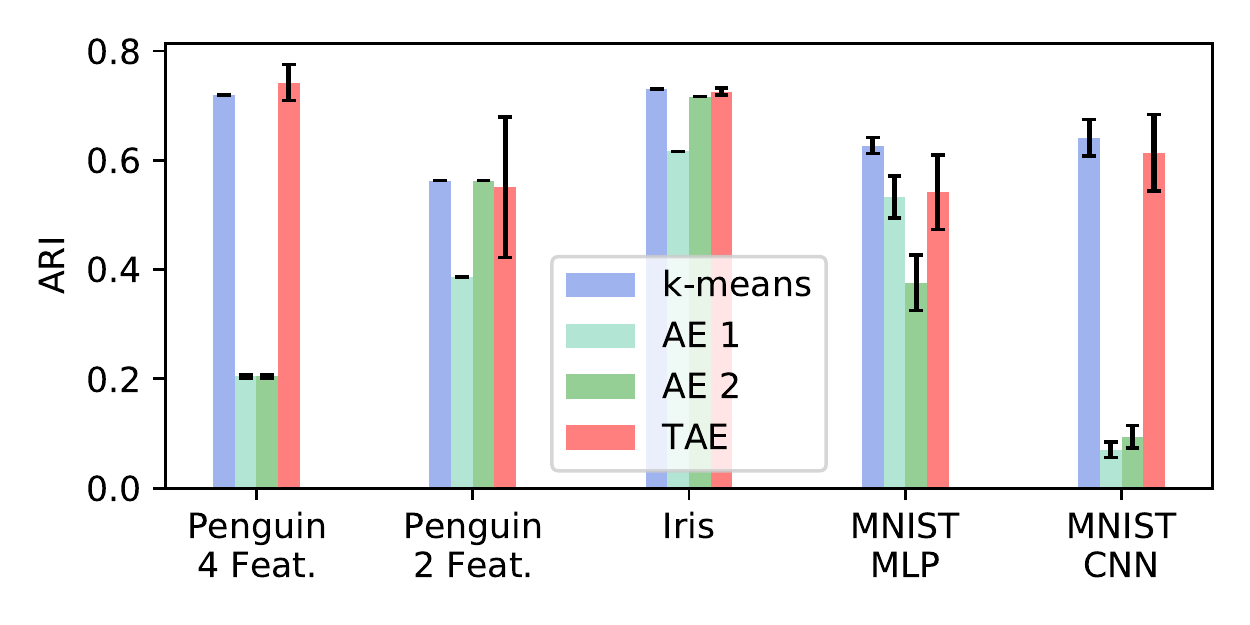}
    \caption{Comparison of $k$-means ++, standard AE with $k$-means ++ and TAE. AE 1 has latent dimension $d$ while AE 2 has latent dimension $d\times k$. MNIST MLP shows the performance for a single hidden layer network, while MNIST CNN considers an encoder and decoder with two convolutional layers each. Again the latent dimension is $d$ for both networks.}
    \label{fig:real clustering}
\end{figure}

To start the analysis we consider the penguin dataset presented in the introduction.
Figure~\ref{fig:penugin scatter} (top left) shows the simpson's paradox for the penguin dataset and the other plots show the clusterings the different algorithms converge to. Notably we see that for $k$-means as well as for both AE architectures, the y-axis parallel decision boundaries are obtained where as for the TAE we obtain clusters that are closer to the true structures.

To further quantify the empirical performance of TAE we show the average and standard deviation over five runs for the 'penguin dataset' with two and four features as well as the 'iris dataset' \cite{Fisher1936Iris1, Iris2} and 'MNIST' \cite{mnist} (sub-sample of $200$ datapoints for class $\{1,2,3,4,5\}$ each) for simple linear networks as well as two layer convolutional networks. We show this in Figure~\ref{fig:real clustering}. For penguin dataset and iris we consider $d=1$ and $d=10$ for MNIST. We observe that the difference in the performance between AE and TAE is significant for the penguin dataset which we attribute to the above outlined clustering structures. For MNIST, we observe very similar performance of AE and TAE, which is most likely due to the fact that the underlying clustering structures have very similar properties, such there is no significant difference for the TAE to exploit.

\subsection{De-Noising on Real Data} \label{sec: de-noising real}
\begin{figure}
    \centering
    \includegraphics[width = 0.6\textwidth]{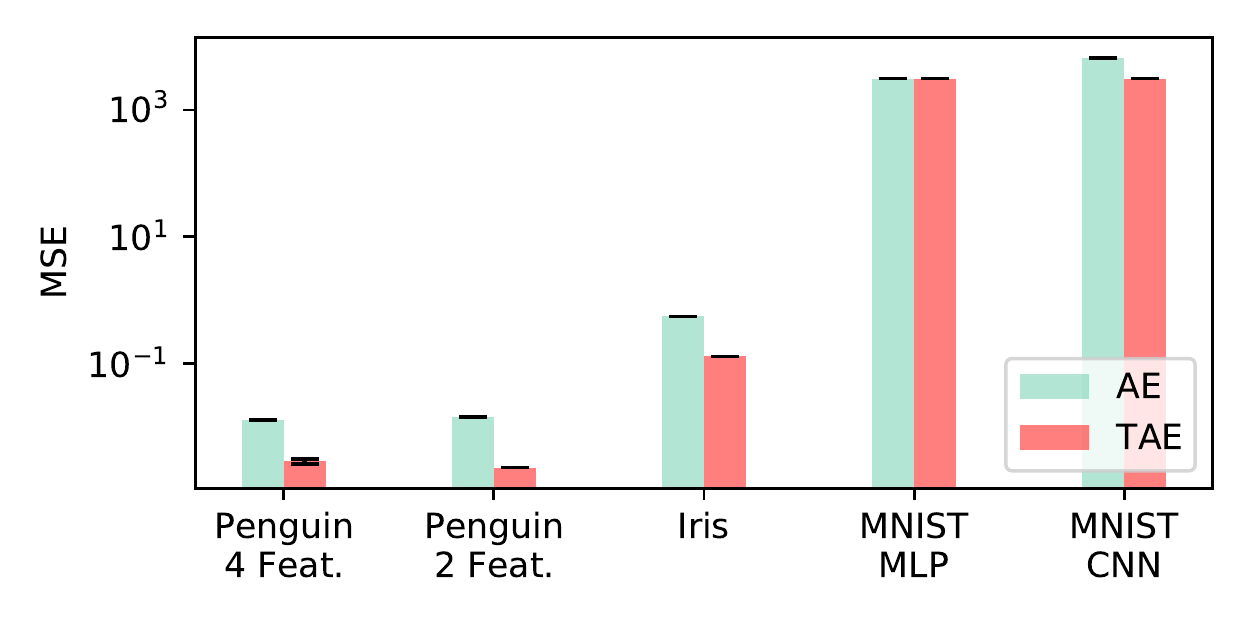}
    \caption{Real data de-noising. Note that the MSE is plotted in log scale }
    \label{fig:real denoising}
\end{figure}

Similar to the experiments on clustering real data we  look at de-noising on various real datasets with standard AE and TAE. We use the same datasets we used in the clustering case to illustrate that TAE learns a reasonable clustering as well as a better reconstruction. The original data was corrupted by adding an isometric Gaussian to each data point.
We illustrate this difference in Figure~\ref{fig:real denoising} and observe that TAE consistently performs the same or significantly better then the standard AE. For the penguin dataset we observe a better cluster recovery and a better de-noising performance. Interestingly for the Iris dataset the reconstruction obtained by TAE is significantly better the the one by the standard AE while their clustering performance is about the same. Finally for MNIST with CNN we get a markedly improved performance on using TAE, while using a single layer linear neural network as the underlying architecture doesn't gives us any significant improvement in performance. This suggests that it is important to choose a function class or underlying architecture that is capable of capturing a low dimensional representation of the data.  

\section{CONNECTION TO EXPECTATION MAXIMIZATION
}\label{sec: EM}
The astute reader would notice that the gradient descent step that we are proposing is similar to Expectation Maximization (EM) algorithm. To explain this connection more carefully (and propose a slightly modified algorithm in the case of Gaussian data), we first write down the EM algorithm itself.

Let $\bX_1, \ldots, \bX_n$ be data coming from from some distribution in the set of distributions with parameter $\theta$ with some latent or unobserved variables $Z$. Let $L(\theta; \bX,Z)$ be the likelihood function of the parameters. 
The EM algorithm then seeks to maximise
\begin{align*}
    Q(\theta) = \sum_{i=1}^n \frac{1}{n}\bbE_{Z|\bX_i,\theta}[\log L(\theta; \bX_i,Z)].
\end{align*}
To make matters a little clearer let us see what this implies when the data comes from a mixture of Gaussians. Let \begin{align*}
    \bX \sim \sum_{i=1}^k p_i N(\mu_i,\Sigma_i).
\end{align*} To imagine a latent variable imagine the data being generated as follows. First sample a random variable $Z$ such that $p(Z=i) = p_i$. Then sample an observation from $N(\mu_z,\Sigma_Z)$. Thus in this case the latent variable $Z$ is the assignment of the data to its respective cluster or the observed data's label. The parameters $\theta$ in this case is the vector $(p_1, ... , p_k, \mu_1, ..., \mu_k, \Sigma_1, ..., \Sigma_k)$. Then given data $\bX_i$, the probability the corresponding $Z$ is $j$ is given by (using Bayes rule) : 
\begin{align*}
    \bS_{j,i} &= \frac{p_j f(\bX_i; \mu_j, \Sigma_j)}{\sum_t p_t f(\bX_i; \mu_t, \Sigma_t)} \\
    &=\frac{p_j \sqrt{\det(2 \pi\Sigma_j)}\exp\left({-\frac{1}{2}(\bX_i-\mu_j)\Sigma_j^{-1}(\bX_i-\mu_j)}\right)}{\sum_t p_t \sqrt{\det(2\pi\Sigma_t)}\exp\left({-\frac{1}{2}(\bX_i-\mu_t)\Sigma_t^{-1}(\bX_i-\mu_t)}\right)}.
\end{align*}
Then the expectation of the log likelihood $\bbE_Z[L(\theta;X,Z)] $ is :
\begin{align*}
    \frac{1}{n}\sum_{i=1}^n \sum_{j=1}^k S_{j,i} \left(\frac{\log \det(2\pi\Sigma_j)}{2}  - \frac{||\Sigma_j^{-1/2}(\bX_i-\mu_j)||^2}{2}\right).
\end{align*}

Now note that ignoring the determinant term ($\log \det(2\pi\Sigma_j)$) maximizing this above quantity is exactly minimizing 
\begin{align*}
    \sum_{i=1}^n \sum_{j=1}^k \bS_{j,i} \norm{\Sigma_j^{-1/2}(\bX_i-\mu_j)}^2.
\end{align*}
Note then  that this is very similar to the term we are optimizing if we allow $S$ to be independent of $\Sigma_j$ and set 
\begin{align*}
\Sigma_j^{-1} = (\bbI - \bV_j\bU_j)^T(\bbI - \bV_j\bU_j) + \lambda \bU_j^T\bU_j,\text{ ~ and ~ }\mu_j = \bC_j.    
\end{align*}
In other words we are fixing the covariance matrix to have numerically low rank, i.e. it has only $k$ (few) of its eigenvalues are $1/\lambda$ (very large) whereas rest of them are $1$ (small).

The connection between TAE and the EM algorithm is especially interesting as there is a vast literature on the theoretical properties of EM and therefore opens up future research directions for a more fine-grained analysis of TAEs.

\section{RELATED WORK AND FUTURE APPLICATIONS}\label{sec: related work}

\textbf{Theoretical analysis of AE.} The theoretical understanding of simple AE is still limited and mainly summarized in Theorem~\ref{Th: linear AE optimal} and formalizes the optimal parameterization of linear AE depending on the considered regularization \cite{BALDI1989NN, Kunin20219pmlr_AE_losss_landscape, Bao2020Neurips_AE_PC, Pretorius2018LearningDO}. While the considered proof techniques differ in our work we derive a similar result of the TAE. An important future direction for theoretical analysis for both simple AE and TAE is the extension to the non-linear setting.

Before going into the related work on clustering and de-noising we would like to note here that the focus of this work is to show the difference in learning a single representation for the data and a representation for each cluster in the data. The general literature of possible, task specific, AE models is vast and would exceed the limits of this related work section. Therefore we focus on the most relevant related work, which is the basic setup for clustering on embedding and de-noising with simple AEs.

\textbf{Clustering.} The main goal of clustering is to group similar objects into the same class in an unsupervised setting. While this problem has been extensively studied in traditional machine learning in terms of 
feature selection \cite{Boutsidis2009NIPS_kmeans_feature_selection,Alelyani2013Book_FeatureSF},
distance functions \cite{Eric200NIPS_Distance_metric_sideinformation,Xiang2008PR_Learning_Mahalanobis} and 
group methods \cite{Macqueen67somemethods,Luxburg2007Book_SpectralClustering,Tao2004ICML_Entropy_Clustering}
(for a more comprehensive overview see \cite{Charu2014Book_Data_clustering})
the time complexity significantly increases with high dimensional data, previous work focuses on projecting data into low-dimensional spaces and then cluster the embedded representations \cite{Volker2003_NeurIPS_feature_selection,Fei2014AAAI_graph_representation_clustering,Zhangyang2016SIAM_Architecture_clustering}. 
For there there several methods have been developed that use deep unsupervised models to learn representations with a clustering focus that simultaneously learns feature representations and cluster assignments using deep neural networks \cite{Xie2016ICML_deep_embedding,Kamran2017_ICCV_Joint_AE,Zhangyang2016SIAM_Architecture_clustering,Xie2015IntegratingIC,Zhangyang2015ICAI_Discriminative_Clustering}. 

\textbf{De-noising.} We consider de-noising with AEs \cite{Antoni2005SIAP_denoising_survay,Cho2013pmlr_DAE}. While there are several extensions to more complex AE models and task specific setups (see e.g. \cite{Chaoning2022Survay}) in this work we focus on the question if learning cluster specific representations is beneficial for reducing the reconstruction error, which to the best of our Knowledge has been considered so far.

\textbf{Possible future applications.} We note that while in this paper we focus on \emph{clustering and de-noising} the general concept can be extended to other AE based downstream tasks such as
anomaly detection \cite{Forero2019IEEM_SSL_anomaly, Mayu2014MLSDA_anomaly_nonlinear, Chong2017KDD_anomaly_robust},
image compression \cite{Theis2017ICLR_Lossy_compression, Balle17ICLRendtonedcompression},
super resolution \cite{Kun2017IEEE_coupled_superresolution, Tzu2017IEEE_superres_cuvaturegaussian}
and machine translation \cite{cho2014SSST_MLtranslation, Sutskever2014NIPSSequenceTS}.

Such extensions are especially of interest for future applications as  Figure~\ref{fig: toy data} and Section~\ref{sec: de-noising real} show that TAE outperform standard AE especially when measured in the reconstruction quality. This indicates that tasks such as image compression can benefit from using TAE, as for such tasks the matching to the true clustering is not required.

\section{CONCLUSION}\label{sec: conclusion}

This work presents a meta algorithm which can be used to better adapt any existing autoencoder architecture to datasets in which one might anticipate cluster structures. By jointly learning the cluster structure and low dimensional representations of clusters, the proposed tensor auto-encoder (TAE) directly improves upon Kmeans, applied to a dataset or to an encoding of the same generated by an AE. More importantly, in the context of de-noising or downstream learning tasks, while it is trivial to note that mathematically a TAE can never have worse MSE than a corresponding AE, we verify the same experimentally. On the surface the difference in performance might simply seem to be a matter of more parameters. However we show experimentally (see experiment on real data clustering, Figure~\ref{fig:real clustering}) that even with the same total number of parameters TAE  clusters better and hence gives a more accurate reconstruction than an AE. An open question in this regard would be to show such a result mathematically under certain data assumptions. Another open implementation problem would be to design a more a efficient gradient step for the TAE.

\section{SOCIAL IMPACT}
Usage of traditional autoencoders on large diverse communities often favour creating a single latent monolithic representations mostly representing a single majority. This might create situations where the interests of various smaller communities are ignored. An instance of this might be to use autoencoders to find a couple of parameters which are the most significant markers of a particular disease. Using a tensorized autoencoders might be of interest in these situations. As this might create multiple representations for each community.

% \section{Acknowledgement}
\section*{ACKNOWLEDGEMENT}

This work has been supported by the German Research Foundation through the SPP-2298 (project GH-257/2-1),  GRK 2428, and also jointly with French National Research Agency through the DFG-ANR PRCI ASCAI.

\clearpage
\printbibliography
\clearpage

\appendix

\section{Additional Illustrations}\label{sec: additional illustration}

\begin{figure}[h]
         \centering
          \includegraphics[width = 0.8\textwidth]{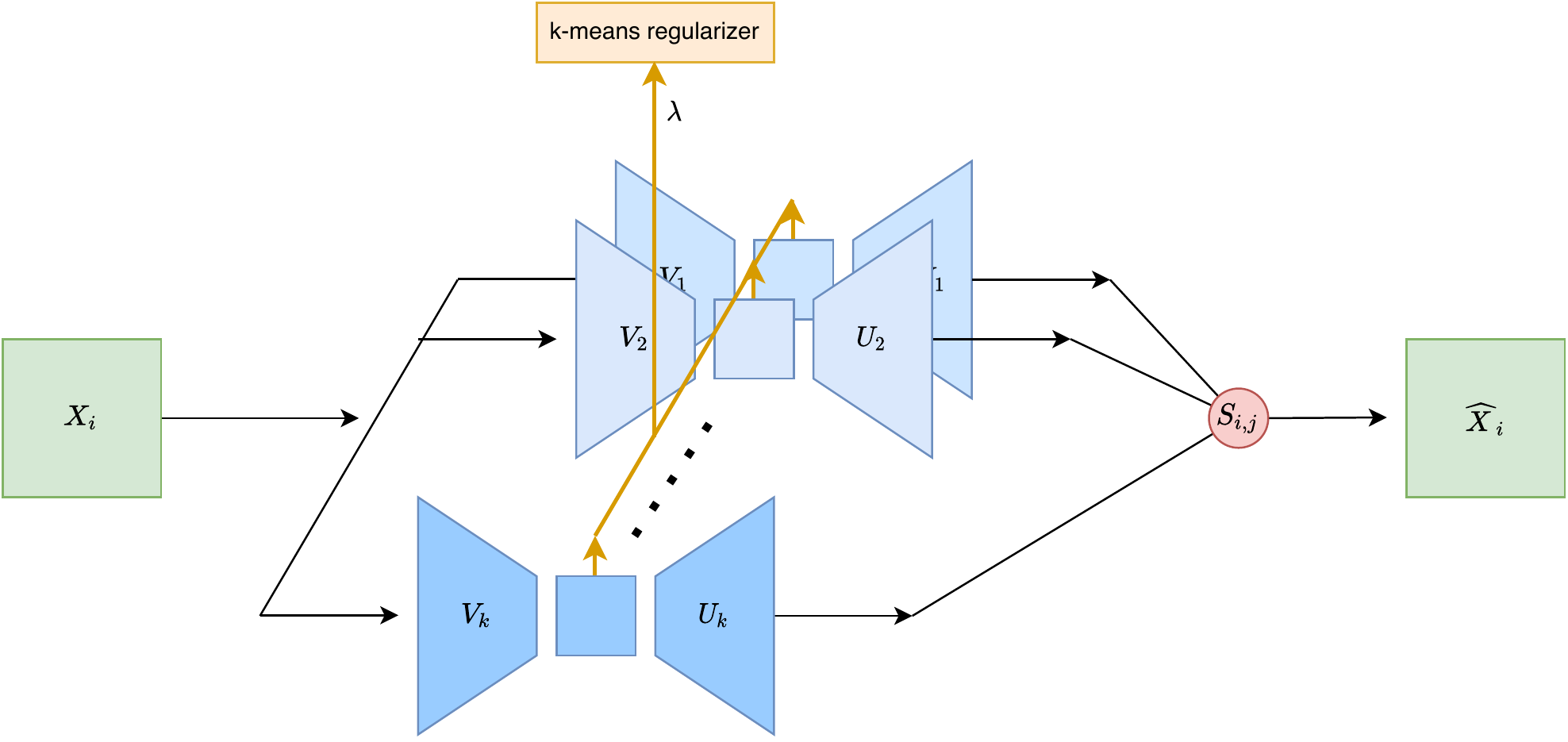}
    \caption[b]{Our approach: Illustration of the tensorized AE. Consider an input $\bX_i$, will be passed through $k$ AE, with $k$-means regularized latent space that reconstruct $\bX_i$. $\bS_{i,j}$ weight the outputs of the AEs by class assignment.}
         \label{fig: TAE illustration}
     \end{figure}

The following (Figure~\ref{fig: TAE illustration}) illustrates the general concept of tensorized k-means regularized AE.
     
\clearpage

\clearpage
\section{Proof of Theorem 2.2}\label{sec: relaxed cost function}
For clarity let us first restate the considered loss function
 \begin{align}\label{eq: relaxed cost supp}
     \loss_\lambda(\bX) := &\sum^n_{i=1} \sum^k_{j = 1}\bS_{j,i}\Big[\norm{\left(\bX_i - \bC_j\right) - \bV_j\bU_j\left(\bX_i - \bC_j\right)}^2 \nonumber\\
     &~ + ~ \lambda \norm{\bU_j\left(\bX_i - \bC_j\right)}^2\Big], \nonumber\\
      \nonumber\\
     \text{s.t. ~ }& \1_k^T\bS = \1_n^T,\quad \bS_{j,i} \geq 0,
 \end{align}
where we define $\bS$ to be a $k \times n$ matrix, such that $\bS_{j,i}$ is the probability that datapoint $i$ belongs to class $j$
and the theorem :

\noindent\textbf{Theorem 2.2} (Parameterization at Optimal for TAE)
\textit{
For $0 < \lambda \leq 1$, optimizing Eq.~\ref{eq: relaxed cost supp} results in the parameters at the optimum satisfying the following:
\begin{enumerate}
   \item[i)]  \textbf{Class Assignment.} While in Eq.~\ref{eq: relaxed cost supp} we define $\bS_{j,i}$ as the probability that $\bX_i$ belongs to class $j$ at the optimal $\bS_{j,i} = 1 \textit{ or } 0$ and therefore converges to a strict class assignment.
    \item[ii)] \textbf{Centers.} $\bC_j$ at optimum naturally satisfies the condition 
    \begin{align*}
    \bC_j = \frac{\sum_i \bS_{j,i}\bX_i}{\sum_i \bS_{j,i}}.
    \end{align*} 
    \item[iii)] \textbf{Encoding / Decoding (learned weights).}  We first show that $\bV_j^T = \bU_j$, and define 
    \[ 
    \hat{\mathbf{\Sigma}}_j: =\sum^n_{i=1}\bS_{j,i}\left(\bX_i - \bC_j\right)\left(\bX_i - \bC_j\right)^T,
    \] 
    then the encoding corresponds to the top $k$ eigenvectors of $\hat{\Sigma}_j$.
\end{enumerate}
}

\begin{proof}
(i) This part is trivial once we note that for a fixed $\bC_j, \bU_j, \bV_j$, the loss is linear in $\bS_{j,i}$. Thus the optimal must occur at the extreme points of the constraints \begin{align*}
    \1_k^T\bS &= \1_n^T \\
    \bS_{j,i} &\geq 0.
\end{align*}

Let us first prove part (iii). To get conditions on $\bU_j, \bV_j$ let us fix $\bS, \bC_j$ and get conditions of optimal in terms of the fixed quantities. Define 
\[
\hat{\Sigma}_j: =\sum^n_{i=1} \bS_{j,i}(\bX_i - \bC_j)(\bX_i - \bC_j)^T
\]
Let us now optimize $\bV_{j},\bU_{j}$ for each fixed $j$.
\begin{align*}
&\min_{\bV_{j},\bU_{j}}\sum^n_{i=1} \bS_{j,i}\left(\norm{(\bX_i - \bC_j) - \bV_j\bU_j(\bX_i - \bC_j)}^2 + \lambda\norm{\bU_j(\bX_i - \bC_j)}^2\right)    \\
=&\min_{\bV_{j},\bU_{j}}\sum^n_{i=1} \bS_{j,i}\left(\norm{(\bbI - \bV_j\bU_j)(\bX_i - \bC_j)}^2 + \lambda\norm{\bU_j(\bX_i - \bC_j)}^2\right) \\
=&\min_{\bV_{j},\bU_{j}}\sum^n_{i=1} \bS_{j,i}\left((\bX_i - \bC_j)^T(\bbI - \bV_j\bU_j)^T(\bbI - \bV_j\bU_j)(\bX_i - \bC_j) + \lambda(\bX_i - \bC_j)^T\bU_j^T\bU_j(\bX_i - \bC_j)\right) \\
=&\min_{\bV_{j},\bU_{j}}\sum^n_{i=1} \bS_{j,i}\Tr\left[(\bX_i - \bC_j)^T(\bbI - \bV_j\bU_j)^T(\bbI - \bV_j\bU_j)(\bX_i - \bC_j)\right] + \lambda\be_j^T\bS\be_i\Tr\left[(\bX_i - \bC_j)^T\bU_j^T\bU_j(\bX_i - \bC_j)\right]\\
=&\min_{\bV_{j},\bU_{j}}\sum^n_{i=1} \bS_{j,i}\Tr\left[(\bbI - \bV_j\bU_j)^T(\bbI - \bV_j\bU_j)(\bX_i - \bC_j)(\bX_i - \bC_j)^T\right] + \lambda\be_j^T\bS\be_i\Tr\left[\bU_j^T\bU_j(\bX_i - \bC_j)(\bX_i - \bC_j)^T\right]\\
=&\min_{\bV_{j},\bU_{j}}\Tr\left[(\bbI - \bV_j\bU_j)^T(\bbI - \bV_j\bU_j)\hat{\Sigma}_j\right] + \lambda\Tr\left[\bU_j^T\bU_j\hat{\Sigma}_j\right]\\
=&\min_{\bV_{j},\bU_{j}}\Tr\left[((\bbI - \bV_j\bU_j)^T(\bbI - \bV_j\bU_j) + \lambda\bU_j^T\bU_j)\hat{\Sigma}_j\right]\\
=&\min_{\bV_{j},\bU_{j}}\Tr\left[(\bbI - \bV_j\bU_j - \bU_j^T\bV_j^T + (1+\lambda)\bU_j^T\bU_j)\hat{\Sigma}_j\right]
\end{align*}
Optimize over $\bV_j$:
\begin{align*}
    \min_{\bV_{j}} -\Tr\left[(\bU_{j}^T\bV_{j}^T  + \bV_{j}\bU_{j}) \hat{\Sigma}_j \right] &= \max_{\bV_{j}} \Tr\left[\bU_{j}^T\bV_{j}^T\hat{\Sigma}_j \right]  + \Tr\left[\bV_{j}\bU_{j}\hat{\Sigma}_j \right] \\
    &= \max_{\bV_{j}} \Tr\left[\bV_{j}^T\hat{\Sigma}_j\bU_{j}^T \right]  + \Tr\left[\bU_{j}\hat{\Sigma}_j\bV_{j} \right] \\
    &= 2\max_{\bV_{j}} \Tr\left[\bU_{j}\hat{\Sigma}_j\bV_{j}\right] \\
    &= 2\max_{\bV_{j}} \Tr\left[\left(\bU_{j}\hat{\Sigma}_j^{1/2}\right)\left(\bV_{j}^T\hat{\Sigma}_j^{1/2}\right)^T\right].
\end{align*}
As trace forms a proper norm, the above has a global maximum at $\bV_{j}^T = \bU_{j}$. Plugging this back into our original problem then we get :
\begin{align*}
&\min_{\bV_{j},\bU_{j}}\sum^k_{j = 1}\sum^n_{i=1} \bS_{j,i}\left(\norm{(\bX_i - \bC_j) - \bV_j\bU_j(\bX_i - \bC_j)}^2 + \lambda\norm{\bU_j(\bX_i - \bC_j)}^2\right)\\
=&\min_{\bV_{j},\bU_{j}}\Tr\left[(\bbI - \bV_j\bU_j - \bU_j^T\bV_j^T + (1+\lambda)\bU_j^T\bU_j)\hat{\Sigma}_j\right]\\
=&\min_{\bU_{j}}\sum^k_{j = 1} \Tr\left[\left(\bbI - (1-\lambda)\bU^T_j\bU_j\right)\hat{\Sigma}_j\right] \\
=& \sum^k_{j = 1} \Tr\left[\hat{\Sigma}_j\right] - (1-\lambda) \max_{\bU_{j},\bS}\sum^k_{j = 1} \Tr\left[\bU^T_j\bU_j\hat{\Sigma}_j\right] & \textit{since } \lambda \leq 1
\end{align*}
Using the orthonormality of $\bU_{j}$ now immediately gives that its collumns must form the top $h$ eigenvectors of $\hat{\Sigma}_j$.

Let us finally derive (ii). Since at optimality now we know that $\bV_j = \bU_j^T$, we can plug that in. As in the previous section we instead fix $\bU_j, \bS$ and get conditions of optimality for $\bC_j$ in terms of them. We will then derive conditions on $\bC_j$ independent of $\bU_j$
First fix $j$ and define:
\[
\bA:=\left(\bbI - (1-\lambda)\bU^T_j\bU_j\right)
\]
then
\begin{align*}
&\min_{\bC_j}\Tr\left[\left(\bbI - (1-\lambda)\bU^T_j\bU_j\right)\left(\sum^n_{i=1}\bS_{j,i}(\bX_i - \bC_j)(\bX_i - \bC_j)^T\right)\right]\\
=&\min_{\bC_j} \Tr\left[\bA\left(\sum^n_{i=1}\bS_{j,i}(\bX_i - \bC_j)(\bX_i - \bC_j)^T\right)\right]\\
=&\min_{\bC_j} \Tr\left[\bA\left(\sum^n_{i=1}\bS_{j,i}(\bX_i\bX_i^T - \bX_i\bC_j^T + \bC_j\bC_j^T)\right)\right]\\
=&\min_{\bC_j} \Tr\left[\bA\left(\left(\sum^n_{i=1}\bS_{j,i}\bX_i\bX_i^T\right) - \left(\sum^n_{i=1}\bS_{j,i}\bX_i\right)\bC_j^T + \left(\sum^n_{i=1}\bS_{j,i}\right)\bC_j\bC_j^T\right)\right].
\end{align*}
derive by $\bC_j$ and set to zero
\begin{align*}
    \frac{\partial}{\partial\bC_j} = -\bA\left(\sum^n_{i=1}\bS_{j,i}\bX_i\right) + \bA\left(\sum^n_{i=1}\bS_{j,i}\right)\bC_j = 0
\end{align*}

Finally we note that as $\bU_j$ is orthonormal, the eigenvalues of $\bU_j\bU_j$ are $1$ and $0$. Thus if $\lambda$ is not zero, $A$ is invertible (as its eigenvalues are $1$ and $\lambda$). Therefore the above expression is minimized for
\begin{align*}
    \bC_j &= \frac{\sum_{i=1}^n \bS_{j,i} \bX_i}{\sum_{i=1}^n \bS_{j,i}}.
\end{align*}

\end{proof}

\clearpage

\section{Experimental Details for Results from Main Paper}\label{sec: experimental details main paper}

In this section we provide the setup for the experiments performed in the main paper. Further details can be see in the provided code.

\subsection{Experiments on Toy Data}
For all experiments we consider the theoretical setting where we use a two layer linear network and $\lambda = 0.1$.
The Scatter plot is for a converged clustering from a randomly sampled run.
The \emph{cluster accuracy} and \emph{de-nosing error} plot show the average over five random initialization after 100 epochs.

We consider $h=1$

\subsection{Notes on the penguin dataset}

We consider two main versions of the dataset. The \emph{two feature} version only considers 'culmen depth' and 'culmen length' as features (this version is illustrated in Figure~1 (main paper)). The \emph{four feature} version additionally considers 'flipper length' in mm and  'body mass' in grams.

Since the features 'body mass' and 'culmen depth' are on very different scales we normalize all features for the experiments with four features.

\subsection{Experiments on Clustering on Real Data}

As a default network we consider a one hidden layer, ReLU network. For Penguin and Iris we consider $h = 1$ and for MNIST $h = 10$.

For the CNN we consider a network with two encoding CNN layers and two decoding CNN layers and the same hidden dimension as for the fully connected setting. For the exact architecture we refer to the provided code.

\subsection{Experiments on De-Noising on Real Data}

Finally for additional illustration of the de-noising results we can consider Figure~\ref{fig:real denoising}, which shows the same results as Figure~5 (main paper) but now normalized by the maximum value and not on a log scale. This further illustrates the significant difference between reconstruction of AE and TAE.

We generally consider the same setup as for the clustering tasks with a default network of one hidden layer, ReLU network. For Penguin and Iris we consider $h = 1$ and for MNIST $h = 10$.

\begin{figure}[b]
    \centering
    \includegraphics[width = 0.6\textwidth]{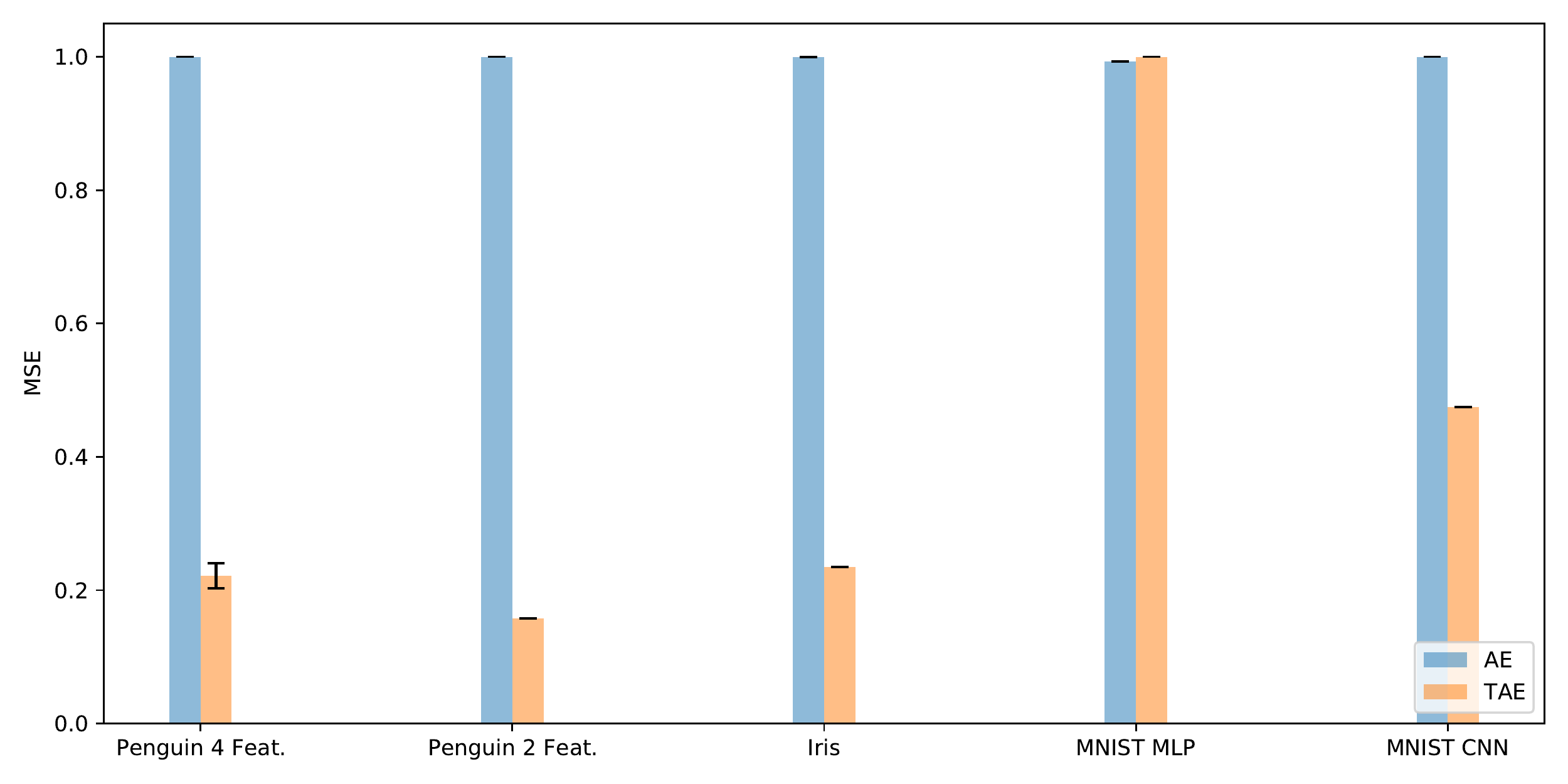}
    \caption{Real data denoising. Plotted is the MSE, normalized on the max for each of the datasets.}
    % \label{fig:real denoising}
\end{figure}

\end{document}